\documentclass[11pt,a4paper]{article}
\usepackage[hyperref]{eacl2021}
\usepackage{times}
\usepackage{latexsym}

\usepackage{multirow}
\usepackage{multicol}
\usepackage{amsmath}
\usepackage{helvet}
\usepackage{courier}
\usepackage{booktabs}
\usepackage{graphicx}
\usepackage{array} 
\usepackage{color}
\usepackage{multirow}
\usepackage[normalem]{ulem}
\aclfinalcopy

\newcolumntype{R}[1]{>{\raggedleft\arraybackslash}p{#1}}
\newcolumntype{L}[1]{>{\raggedright\arraybackslash}p{#1}}
\newcolumntype{C}[1]{>{\centering\arraybackslash}p{#1}}

\usepackage{microtype}

\title{Joint Coreference Resolution and Character Linking \\ for Multiparty Conversation}

\author{Jiaxin Bai$^1$, Hongming Zhang$^1$, Yangqiu Song$^1$, and Kun Xu$^2$ \\
  $^1$CSE, HKUST \\
  $^2$ Tencent AI Lab \\
  \texttt{\{jbai, hzhangal, yqsong\}@cse.ust.hk, kxkunxu@tencent.com} \\}

\date{}

\begin{document}
\maketitle
\begin{abstract}
Character linking, the task of linking mentioned people in conversations to the real world, is crucial for understanding the conversations.
For the efficiency of communication, humans often choose to use pronouns (e.g., ``she'') or normal phrases  (e.g., ``that girl'') rather than named entities (e.g., ``Rachel'') in the spoken language, which makes linking those mentions to real people a much more challenging than a regular entity linking task.
To address this challenge, we propose to incorporate 
the richer context from the coreference relations
among different mentions to help the linking.
On the other hand, considering that finding coreference clusters itself is not a trivial task and could benefit from the global character information, we propose to jointly solve these two tasks.
Specifically, we propose C$^2$, the joint learning model of \textbf{C}oreference resolution and \textbf{C}haracter linking. 
The experimental results demonstrate that C$^2$ can significantly outperform previous works on both tasks.
Further analyses are conducted to analyze the contribution of all modules in the proposed model and the effect of all hyper-parameters.

\end{abstract}

\section{Introduction}

Understanding conversations has long been one of the ultimate goals of the natural language processing community, and a critical step towards that is grounding all mentioned people to the real world.
If we can achieve that, we can leverage our knowledge about these people (e.g., things that happened to them before) to better understand the conversation.
On the other hand, we can also aggregate the conversation information back to our understanding about these people, which can be used for understanding future conversations that involve the same people.
To simulate the real conversations and investigate the possibility for models to ground mentioned people, the character linking task was proposed~\cite{chen-choi-2016-character}.
Specifically, it uses the transcripts of TV shows (i.e., \textit{Friends}) as the conversations and asks the models to ground all person mentions to characters.

\begin{figure}[t]
\begin{center}
\includegraphics[clip, trim=3cm 13cm 15cm 4cm,width=0.45\textwidth]{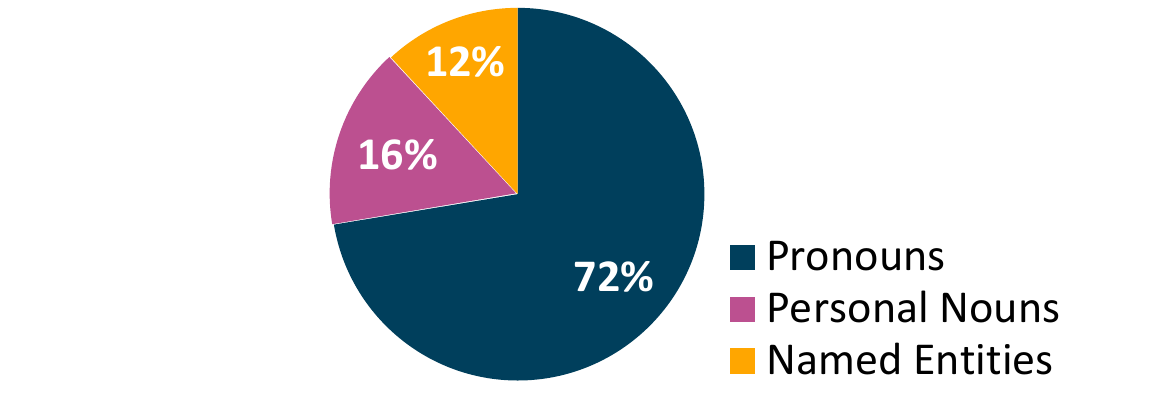}
\end{center}

\caption{
The composition of the mentions in conversations for character grounding.
Over 88\% of the mentions are not named entities, which brings exceptional challenges when linking those to character entities. 
}
\label{fig:mentionCount}
\end{figure}

Even though the character linking task can be viewed as a special case of the entity linking task, it is more challenging than the ordinary entity linking task for various reasons.
First, the ordinary entity linking task often aims at linking named entities to external knowledge bases such as Wikipedia, where rich information (e.g., definitions) are available.
However, for the character linking task, we do not have the support of such rich knowledge base and all we have are the names of these characters and simple properties (e.g., gender) about these characters.
Second, the mentions in the ordinary entity linking are mostly concepts and entities, but not pronouns.
However, as shown in Figure \ref{fig:mentionCount}, 88\% of the character mentions are pronouns (e.g., ``he'') or personal nouns (e.g., ``that guy'') while only 12\% are named entities.

\begin{figure}[t]
\begin{center}
\includegraphics[clip, trim=3cm 9cm 2cm 4cm,width=\linewidth]{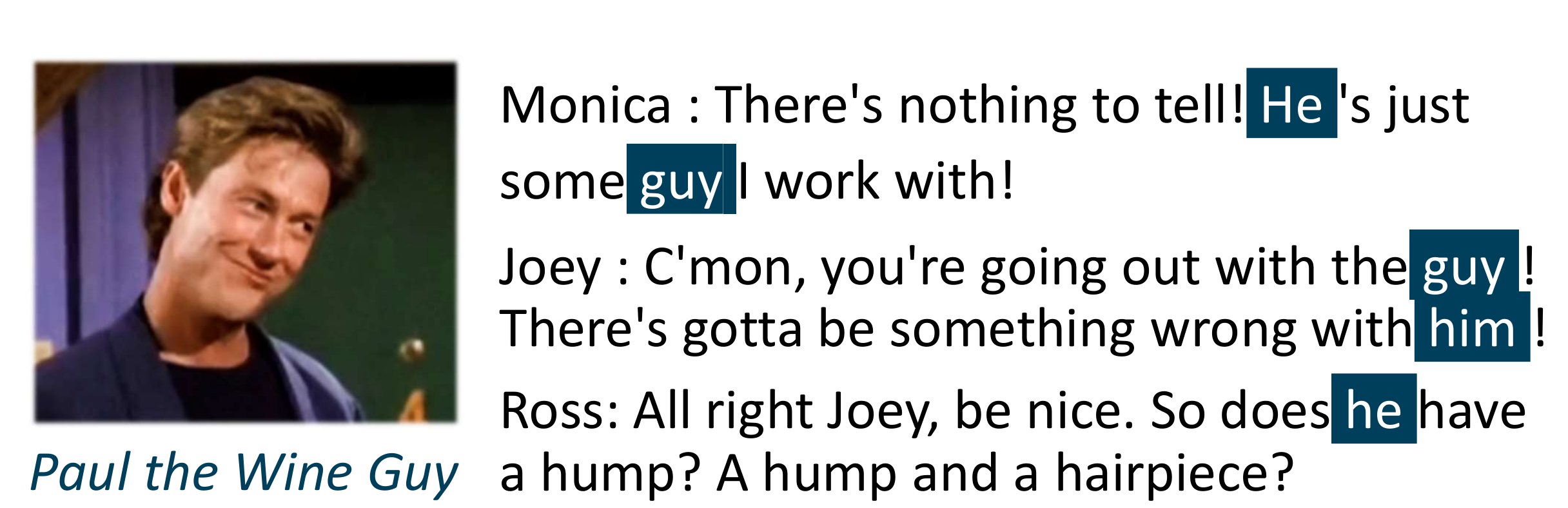}
\end{center}
\caption{
	 Coreference clusters can help to connect the whole conversation to provide a richer context for each mention such that we can better link them to \textit{Paul}. Meanwhile, the character \textit{Pual} can also provide global information to help resolve the coreference.
}
\label{fig:alan}
\end{figure}

Considering that pronouns have relatively weak semantics by themselves, to effectively ground mentions to the correct characters, we need to fully utilize the context information of the whole conversation rather than just the local context they appear in.
One potential solution is using the coreference relations among different mentions as the bridge to connect the richer context.
One example is shown in Figure~\ref{fig:alan}.
It is difficult to directly link the highlighted mentions to the character \textit{Paul} based on their local context 
because the local context of each mention can only provide a single piece of information about its referent, e.g., ``person is a male'' or ``the person works with Monica.'' 
Given the coreference cluster, the mentions refer to the same person, and the pieces of information are put together to jointly determining the referent.
As a result, it is easier for a model to do character linking with resolved coreference.
Similar observations are also made in~\cite{chen-etal-2017-robust}.

At the same time, we also noticed that coreference resolution, especially those involving pronouns, is also not trivial.
As shown by the recent literature on the coreference resolution task \cite{lee-etal-2018-higher,kantor-globerson-2019-coreference}, the task is still challenging for current models and the key challenge is how to utilize the global information about entities.
And that is exactly what the character linking model can provide.
For example, in Figure~\ref{fig:alan}, 
it is difficult for a coreference model to correctly resolve the last mention \textit{he} in the utterance given by \textit{Ross} based on its local context, 
because another major male character (\textit{Joey}) joins the conversation, which can distract and mislead the coreference model.
However, if the model knows the mention \textit{he} links to the character \textit{Paul} and \textit{Paul} works with \textit{Monica},
it is easier to resolve \textit{he} to some \textit{guy} that Monica works with.

Motivated by these observations, we propose to jointly train the \textbf{C}oreference resolution and \textbf{C}haracter linking tasks and name the joint model as C$^2$.
C$^2$ adopts a transformer-based text encoder and includes a mention-level self-attention (MLSA) module that enables the model to do mention-level contextualization.  
Meanwhile, a joint loss function is designed and utilized so that both tasks can be jointly optimized.
The experimental results 
demonstrate that C$^2$ outperforms all previous work significantly on both tasks.
Specifically, compared with the previous work~\cite{zhou-choi-2018-exist}, C$^2$ improves the performance by 15\% and 26\% on the coreference resolution and character linking tasks\footnote{The performance on the coreference resolution is evaluated based on the average F1 score of B3, CEAF$_{\phi}$4, and BLANC. The performance on the character linking task is evaluated by the average F1 score of the micro and macro F1.} respectively  comparing to the previous state-of-the-art model ACNN \cite{zhou-choi-2018-exist} .
Further hyper-parameter and ablation studies testify the effectiveness of different components of C$^2$ and the effect of all hyper-parameters. Our code is available at \url{https://github.com/HKUST-KnowComp/C2}.

\section{Problem Formulations and Notations}

We first introduce the coreference resolution and character linking tasks as well as used notations.
Given a conversation, which contains multiple utterances and $n$ character mentions $c_1, c_2, ..., c_n$, and a pre-defined character set $\mathcal{Z}$, which contains $m$ characters $z_1, z_2, ..., z_m$.
The coreference resolution task is grouping all mentions to clusters such that all mentions in the same cluster refer to the same character. 
The character linking task is linking each mention to its corresponding character.

\section{Model}

\begin{figure}[t]
\begin{center}
\includegraphics[clip, trim=5cm 7.5cm 15cm 2.2cm, width=\linewidth]{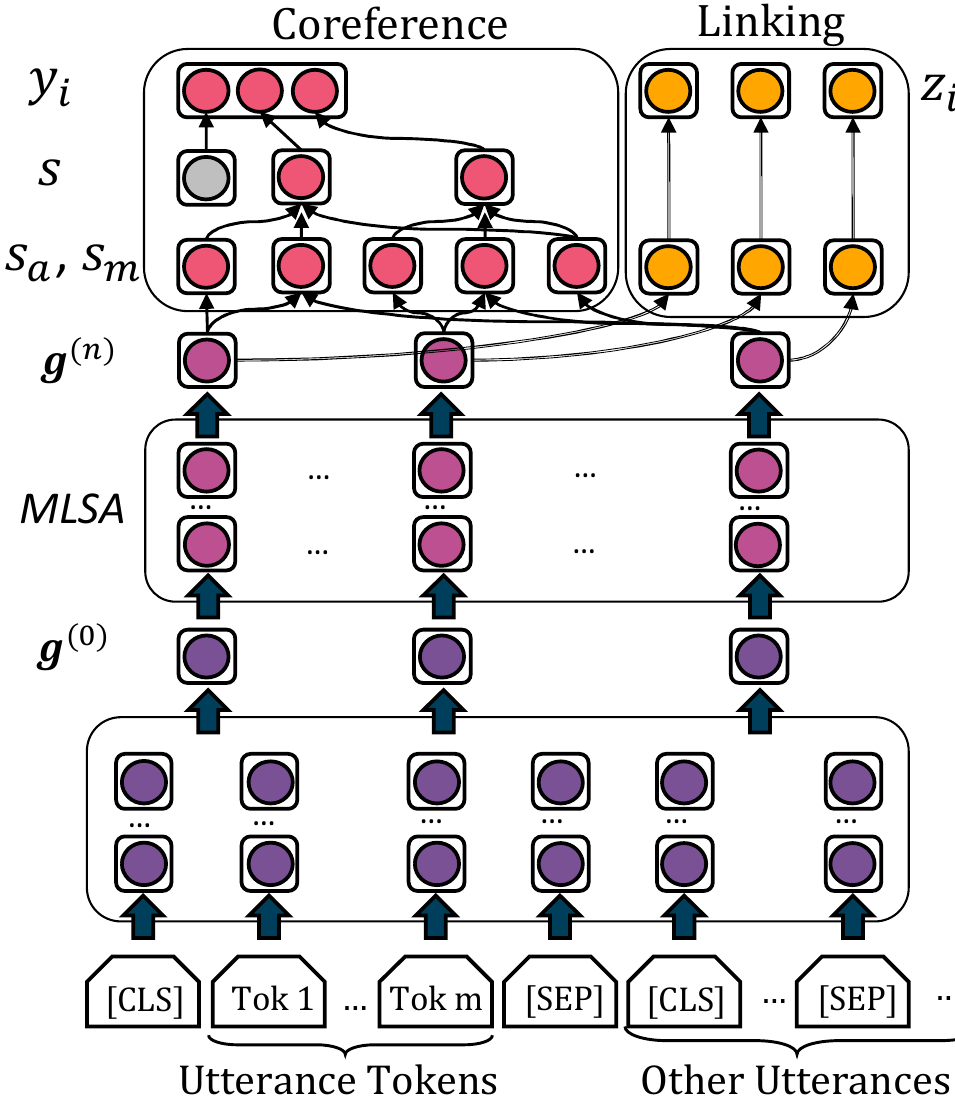}
\end{center}
\caption{
The coreference module and the linking module share the same mention representation $g^{(n)}$ as inputs.
The mention representation $g^{(i)}$ are iteratively refined through the mention-level self-attention layers. 
The initial mention representations $g^{(0)}$ are the sum of text span representations from a pre-trained text encoder and corresponding speaker embeddings.
}
\label{fig:model}
\end{figure}

In this section, we introduce the proposed C$^2$ framework, which is illustrated in Figure~\ref{fig:model}.
With the conversation and all mentions as input, we first encode them with a shared mention representation encoder module, which includes a pre-trained transformer text encoder and a mention-level self-attention (MLSA) module. 
After that, we make predictions for both tasks via two separate modules.
In the end, a joint loss function is devised so that the model can be effectively trained on both tasks simultaneously. 
Details are as follows.

\subsection{Mention Representation}

We use pre-trained language models \cite{devlin2018bert, joshi2019spanbert} to obtain the contextualized representations for mentions.
As speaker information is critical for the conversation understanding, we also include that information by appending speaker embeddings to each mention.
As a result, the initial representation of mention $i$ is:
\begin{equation}
    g_i^{(0)} = t_{start_i} + t_{end_i} + e_{speaker_i},
\end{equation}
where $t_{start_i}$ and $t_{end_i}$ are the contextualized representation of the beginning and the end tokens of mention $i$, and the $e_{speaker_i}$ is the speaker embedding for the current speaker.
Here, we omit the embeddings of inner tokens because their semantics has been effectively encoded via the language model. 
The speaker embeddings are randomly initialized before training.

Sometimes the local context of a mention is not enough to make reasonable predictions, and it is observed that the co-occurred mentions can provide document-level context information.
To refine the mention representations given the presence of other mentions in the document, we introduce the Mention-Level Self-Attention (MLSA) layer, which has $n$ layers of transformer encoder structure \cite{Vaswani2017AttentionIA} and is denoted as $T$.
Formally, this iterative mention refinement process can be described by
\begin{align} 
g_1^{(i+1)},...,g_k^{(i+1)}= T(g_1^{(i)},...,g_k^{(i)}),
\end{align}
where $k$ indicates the number of mentions in a document, and the $g^{(i)}$ means the mention representation from the $i$-th layer of MLSA.

\begin{table*}[t]
\begin{center}
\small
\begin{sc}
\begin{tabular}{p{1.5cm}||C{1.7cm}|C{1.7cm}|C{1.7cm}|C{1.7cm}|C{1.7cm}|C{1.7cm}}

\toprule
 Dataset &	Episodes&	Scenes&	Utterances	&	Speakers &	Mentions &	Entities \\
\midrule

TRN &	76&	987	&18,789	&	265	&36,385	&628 \\
DEV &	8&	122&	2,142&		48&	3,932&	102\\
TST	&13&	192&	3,597&		91&	7,050	&165\\
\midrule
Total 	&97&	1,301&	24,528&		331&	47,367&	781\\
\bottomrule
\end{tabular}
\end{sc}
\end{center}
\caption{The detailed information about the datasets. For each season, the episode 1 to 19 are used for training, the episode 20 to 21 for development, and the remaining for testing.}
\label{tab:Dataset }
\end{table*}

\subsection{Coreference Resolution}

Following the previous work~\cite{joshi2019spanbert}, we model the coreference resolution task as an antecedent finding problem.
For each mention, we aim at finding one of the previous mentions that refer to the same person. 
If no such previous mention exists, it should be linked to the dummy mention $\varepsilon$.
Thus the goal of a coreference model is to learn a distribution, $P(y_i)$ over each antecedent for each mention $i$:
\begin{align}
P(y_i) = \frac{e^{s(i,y_i)}}{\Sigma_{y'\in\mathcal{Y}(i)}e^{s(i,y')}},
\end{align}
where $s(i,j)$ is the score for the antecedent assignment of mention $i$ to $j$.
The score $s(i,j)$ contains two parts: (1) the plausibility score of the mentions $s_a(i,j)$; (2) the mention score measuring the plausibility of being a proper mention $s_m(i)$.
Formally, the $s(i,j)$ can be expressed by
\begin{align}
    s(i, j) &= s_m(i) + s_m(j) + s_a(i, j),\\
    s_m(i) &= FFNN_m(g_i^{(n)}),
    \\ s_a(i,j) &= FFNN_a([g_i^{(n)}, g_j^{(n)}]), \label{eq:sa}
\end{align}
where $g^{(n)}$ stands for the last layer mention representation resulted from the MLSA and $FFNN$ indicates the feed-forward neural network.

\subsection{Character Linking}
The character linking is formulated as a multi-class classification problem, following previous work \cite{zhou-choi-2018-exist}.
Given the mention representations $g^{(n)}$, the linking can be done with a simple feed-forward network, denoted as $FFNN(\cdot)$. 
Specifically, the probability of character entity $z_i$ is linked with a given mention $i$ can be calculated by:
\begin{align}\label{eq:Qzi}
    Q(z_i) = Softmax(FFNN_l(g_i^{(n)}))_{z_i},
\end{align}
where the notation $(.)_z$ represents the $z$-th composition of a given vector. 

\subsection{Joint Learning}
To jointly optimize both coreference resolution and entity linking, we design a joint loss of both tasks.
For coreference resolution, given the gold clusters, we
minimize the negative log-likelihood of the possibility
that each mention is linked to a gold antecedent. 
Then the coreference loss $L_{c}$ 
becomes
\begin{align}
    L_{c} = -\sum_{i=1}^N\log\sum_{y\in \mathcal{Y}(i) \cap GOLD(i)} P(y),
     \label{formula:c_loss}
\end{align}
where the $GOLD(i)$ denotes the gold coreference cluster that mention $i$ belongs to. 
Similarly, for character linking, we minimize the negative log-likelihood of the joint probability for each mention being linked to the correct referent character:
\begin{align}
    L_{l} = -\sum_{i=1}^N \log Q(z_i).
    \label{formula:l_loss}
\end{align}

Finally, the joint loss can be the arithmetic average of the coreference loss and linking loss:
\begin{align}
    L = \frac{1}{2} ( L_{l} + L_{c}).
\end{align}

\section{Experiments}
In this section, we introduce the experimental details to demonstrate the effectiveness of C$^2$.

\begin{table*}[t]
\begin{center}
\scriptsize
\begin{sc}
\begin{tabular}{p{3.5cm}||C{0.6cm}C{0.6cm}C{0.6cm}|C{0.6cm}C{0.6cm}C{0.6cm}|C{0.6cm}C{0.6cm}C{0.6cm}|C{1.6cm}}
\toprule
\multirow{2}{*}{Model} & \multicolumn{3}{c|}{B3} & \multicolumn{3}{c|}{CEAF{$\phi4$} }  & \multicolumn{3}{c|}{BLANC} & \multirow{2}{*}{Ave.F1}\\ 
\cline{2-10}
 & Prec. & Rec. & F1 & Prec.  & Rec. & F1 & Prec. & Rec. & F1  &   \\ 
\midrule

ACNN       &  84.30 &   71.90 &  77.60 & 54.50 &  71.80 &  62.00  & 84.30 & 80.40   & 82.10   & 73.96  (0.97)  \\
CorefQA (SpanBERT-Large)  & 73.72 & 75.55  & 74.62   &  65.82 &  72.38 & 68.94 & 86.82  &  84.69 &  85.75   & 76.44 (0.20)  \\
\midrule
C2F (BERT-Base)     & 69.62 & 76.11   & 72.72  &  66.44 &   60.92 &  63.56 & 79.38  &  86.05 &  82.38   &  72.88   (0.23) \\
C2F (BERT-Large)    & 71.72 & 80.25   & 75.75   &  69.97 &   62.61&  66.08 & 81.65  &  88.23 &  84.63   &  75.49   (0.18) \\
C2F (SpanBERT-Base) & 72.49 & 77.88   & 75.08   &  66.00 &   64.23 &  65.10 & 81.60 &  87.43 &  84.27   & 74.81 (0.19) \\
C2F (SpanBERT-Large) & 81.93 & 84.38   & 82.57   &  78.04 &   71.99 &  74.89 & 88.15  & 91.09 &  89.56  & 82.34 (0.17) \\

\midrule
C$^2$ (BERT-Base) & 78.10 & 81.56   & 79.79  &  72.48 &   69.87 &  71.15 & 86.14  &  89.49 &  87.74   & 80.14 (0.21) \\
C$^2$ (BERT-Large)  & 78.49 & 81.90   & 80.16   &  73.81 &   71.15 &  72.46 & 86.20  &  89.93 &  87.97  & 80.17 (0.23)  \\
C$^2$  (SpanBERT-Base)  & 81.18 & 83.59   & 82.36   &  73.64 &   73.09 &  73.36 & 88.06  &  91.04 &  89.49   & 81.74  (0.19) \\
C$^2$  (SpanBERT-Large) & \textbf{85.83}  & \textbf{85.27 }  & \textbf{85.55 }  &  \textbf{77.13} &   \textbf{77.84 }&  \textbf{77.48} & \textbf{92.31}  &  \textbf{92.03} &  \textbf{92.17} & \textbf{85.06} (0.16) \\
\bottomrule
\end{tabular}
\end{sc}
\end{center}
\caption{Experimental results on the coreference resolution task. The results are presented in a 2-digit decimal following previous work. Standard deviations of the average F1 scores are shown in brackets.
}
\label{tab:Coreference}
\end{table*}

\subsection{Data Description}

\begin{figure}[t]
\begin{center}
\includegraphics[clip, trim=3.2cm 3cm 6.8cm 3cm, width=\linewidth]{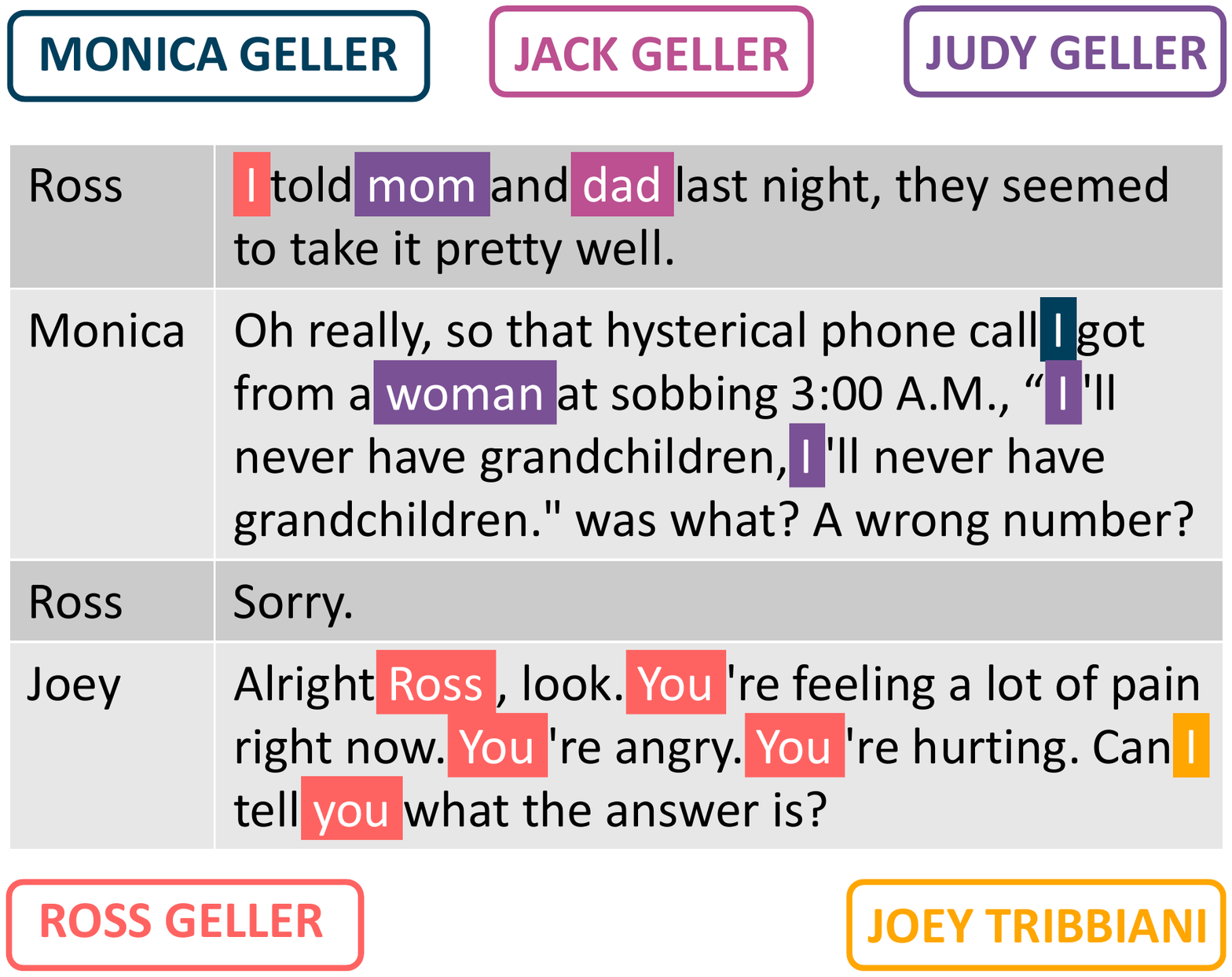}
\end{center}
\caption{
The example annotations for character identification. The arrows in the figure are pointing from the character mentions to their referent character entities.
}
\label{fig:linkingExample}
\end{figure}

We use the latest released character identification V2.0\footnote{https://github.com/emorynlp/character-identification} as the experimental dataset, and we follow the standard training, developing, and testing separation provided by the dataset.
In the dataset, all mentions are annotated with their referent global entities. For example, in Figure~\ref{fig:linkingExample}, 
the mention \textit{I} is assigned to \textit{ROSS}, and the mentions \textit{mom} and \textit{dad} are assigned to \textit{JUDY} and \textit{JACK } respectively in the first utterance given by \textit{Ross}.
 The gold coreference clusters are derived by grouping the mentions assigned to the same character entity. 
Statistically, the dataset includes four seasons of the TV show \textit{Friends}, which contain 97 episodes, 1,301 scenes, and 24,528 utterances. In total, there are 47,367 mentions, which are assigned to 781 unique characters.  The detailed statistics are shown in Table~\ref{tab:Dataset }.

\subsection{Baseline Methods}
The effectiveness of the joint learning model is evaluated on both the coreference resolution and character linking tasks. 
To fairly compare with existing models, only the singular mentions are used following the singular-only setting (S-only) in the previous work \cite{zhou-choi-2018-exist}. 

For the {\bf coreference resolution} task, we compare with the following methods.

\begin{itemize}

\item ACNN: A CNN-based model \cite{zhou-choi-2018-exist} coreference resolution model that can also produce the mention and mention-cluster embeddings at the same time. 

\item C2F: The end-to-end coarse-to-fine coreference model \cite{joshi2019coref} with BERT~\cite{devlin2018bert} or SpanBERT~\cite{joshi2019spanbert} as the encoder.

\item CorefQA: An approach that reformulates the coreference resolution problem as a question answering problem \cite{wu2019coreference} and being able to be benefited from fine-tuned question-answer text encoders.

\end{itemize}

For the {\bf character linking} task, we also include ACNN as a baseline method.
Considering existing general entity linking models~\cite{kolitsas2018endtoend, van_Hulst_2020,raiman2018deeptype, onando2020evaluating} cannot be applied to the character linking problem because they are not designed to handle pronouns, we propose another text-span classification model with transformer encoder as another strong baseline for the character linking task.

\begin{itemize}
\item ACNN: A model that uses the mention and mention-cluster embeddings as input to do character linking \cite{zhou-choi-2018-exist}.
\item BERT/SpanBERT: 
A text-span classification model consists of a transformer text encoder followed by a feed-forward network. 
\end{itemize}

\subsection{Evaluation Metrics}

We follow the previous work \cite{zhou-choi-2018-exist} for the evaluation metrics.
Specifically, for coreference resolution, three evaluation metrics, B$^3$, CEAF$_{\phi4}$, and BLANC,  are used. 
The metrics are all proposed by the CoNNL'12 shared task \cite{Pradhan2012CoNLL}  to evaluate the output coreference cluster against the gold clusters. 
We follow \citet{zhou-choi-2018-exist} to use BLANC \cite{RECASENS2011BLANC} to replace MUC \cite{Vilain1995MUC6} because BLANC takes singletons into consideration but MUC does not.
As for the character linking task, we use the Micro and Macro F1 scores to evaluate the multi-class classification performance.

\subsection{Implementation Details}

In our experiments, we consider four different pre-trained language encoders: BERT-Base, BERT-Large, SpanBERT-Base, and SpanBERT-Large, and we use $n=2$ layers of the mention-level self-attention (MLSA).
The feed-forward networks are implemented by two fully connected layers with ReLU activations. 
Following the previous work, \cite{zhou-choi-2018-exist}, the scene-level setting is used, where, each scene is regarded as a document for coreference resolution and linking.
During the training, each mini-batch consists of segments obtained from a single document. 
The joint learning model is optimized with the Adam optimizer~\cite{DBLP:journals/corr/KingmaB14} with an initial learning rate of 3e-5, and a warming-up rate of 10\%.
The model is set to be trained for 100 epochs with an early stop.
All the experiments are repeated three times, and the average results are reported. 

\begin{table*}[t]
\begin{center}
\scriptsize
\begin{sc}
\begin{tabular}{p{2.8cm}||C{0.7cm}|C{0.7cm}|C{0.7cm}|C{0.7cm}|C{0.7cm}|C{0.7cm}|C{0.7cm}|C{0.7cm}||C{1.4cm}|C{1.4cm}}
\toprule

Model & Ro & Ra & Ch & Mo & Jo & Ph & Em & Ri & Micro& Macro  \\ 
\midrule
ACNN  & 78.3 & 86.5 & 78.8 & 81.7 & 78.3 & 88.8 & 69.2 & \textbf{83.9} & 73.7 (0.6)  & 59.6 (2.3)   \\
\midrule
BERT-Base & 87.4 & 89.9 & 86.6 & 88.2 & 87.1 & 91.1 & 94.3 & 62.4 & 84.0 (0.1)  & 77.3 (0.2)  \\
BERT-Large & 88.2 &  89.9 & 87.9 &  88.8 & 87.7 & 93.1 & 93.5 & 68.0  & 84.8 (0.2)  & 79.1 (0.2) \\
SpanBERT-Base & 87.6 & 91.8 & 86.7 & 88.2 & 86.8 & 92.6 & 94.6 & 73.3 & 84.2 (0.1) &  77.3 (0.2) \\
SpanBERT-Large & 90.9 & 92.8 & 88.3 & 90.3 & 90.2 & 94.3 & 94.6 & 71.7 & 85.5 (0.1) & 79.8 (0.2)  \\
\midrule
C$^2$ (BERT-Base)  & 86.5 & 87.8 & 85.6 & 86.8 & 88.1 & 92.4 & 93.0 & 66.0& 84.0 (0.1)  & 78.6 (0.2) \\
C$^2$ (BERT-Large)  & 85.9 & 90.0 & 87.3 & 86.9 & 87.2 & 93.0 &  \textbf{96.1} & 66.0& 84.9 (0.1)  & 79.5 (0.2)  \\
C$^2$(SpanBERT-Base)  & 89.8 & 91.3 &  90.5 & 90.9 & 87.8 & 93.2 & 93.4 & 71.3  & 85.7 (0.1) & 81.0 (0.1)  \\
C$^2$ (SpanBERT-Large)& \textbf{91.2} &  \textbf{94.1} & \textbf{91.1} & \textbf{92.5} & \textbf{90.4} & \textbf{94.4}  & 89.2 & 77.1 & \textbf{87.0} (0.1)  & \textbf{81.1} (0.1) \\

\bottomrule
\end{tabular}
\end{sc}
\end{center}
\caption{Experimental results per character on the character linking.  The results are presented in a 1-digit decimal following previous work. Standard deviations of the Micro and Macro F1 scores are shown in brackets. The names in the table are written in two-letter acronyms. Ro: Ross, Ra: Rachel, Ch: Chandler, Mo: Monica, Jo: Joey, Ph: Phoebe, Em: Emily, Ri: Richard}
\label{tab:EntityLinkingDetail}
\end{table*}

\section{Results and Analysis}

In this section, we discuss the experimental results and present a detailed analysis.

\subsection{Coreference Resolution Results}

The performances of coreference resolution models are shown in Table \ref{tab:Coreference}. 
C$^2$ with SpanBERT-large achieves the best performance on all evaluation metrics. 
Comparing to the baseline ACNN model, which uses hand-crafted features, C$^2$ uses a transformer to better encode the contextual information.
Besides that, even though ACNN formulates the coreference resolution and character linking tasks in a pipe-line and uses the coreference resolution result to help character linking, the character linking result cannot be used to help to resolve coreference clusters.
As a comparison, we treat both tasks jointly such that they can help each other.

Currently, CorefQA is the best-performing general coreference resolution model on the OntoNotes dataset \cite{Pradhan2012CoNLL}.
However, its performance is limited on the conversation dataset due to two reasons.
First, different from the experimental setting of OntoNotes, the mentions in our experiment setting are gold mentions.
Consequently, the flexible span predicting strategy of CorefQA loses its advantages because of the absence of the mention proposal stage.
Second, the CorefQA leverages the fine-tuning on other question answering (QA) datasets and it is possible that the used QA dataset (i.e., SQuAD-2.0~\cite{DBLP:conf/acl/RajpurkarJL18}) is more similar to OntoNotes rather than the used multiparty conversation dataset, which is typically much more informal. 
As a result, the effect of such fine-tuning process only works on OntoNotes.

The coarse-to-fine (C2F) model \cite{joshi2019coref} 
with a transformer encoder was the previous state-of-the-art model on OntoNotes.  
Referring to Table \ref{tab:Coreference}, given the same text encoder, the proposed C$^2$ model can constantly outperform the C2F model.
These results further demonstrate that with the help of the proposed joint learning framework, the out-of-context character information can help achieve better mention representations so that the coreference models can resolve them more easily.

\subsection{Character Linking Results}

As shown in Table \ref{tab:EntityLinkingDetail}, the proposed joint learning model also achieves the best performance on the character linking task and there are mainly two reasons for that.
First, the contextualized mention representations obtained from pre-trained language encoders can better encode the context information than those representations used in ACNN.
Second, with the help of coreference clusters, richer context about the whole conversation is encoded for each mention.
For example, when using the same pre-trained language model as the encoder, C$^2$ can always outperform the baseline classification model.
These empirical results confirm that, though the BERT and SpanBERT can produce very good vector representation for the mentions based on the local context, the coreference clusters can still provide useful document-level contextual information for linking them to a global character entity.

\begin{figure}[t]
\begin{center}
\includegraphics[width=\linewidth]{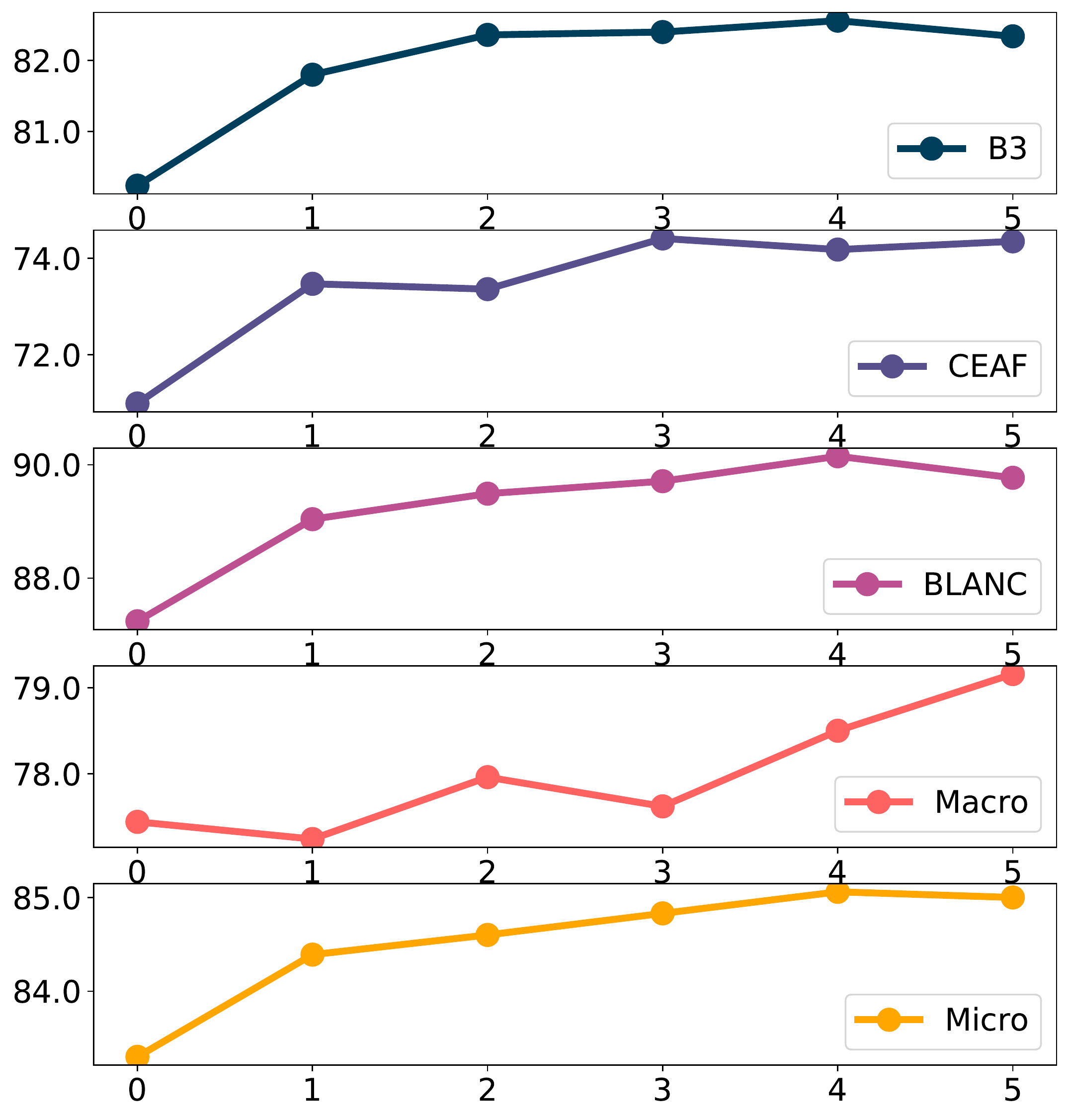}
\end{center}
\caption{
The x-axis is the number of MLSA layers used in the C$^2$. The y-axes are the F1 scores on each metric for their corresponding tasks. The curves have general trends of going up, which indicates that the model performs better when there are more layers. 
}
\label{fig:layergraph}
\end{figure}

\subsection{The Number of MLSA Layers}

\begin{figure*}[t]
\begin{center}
\includegraphics[clip, trim=0.3cm 4cm 3cm 2.5cm, width=0.92\textwidth]{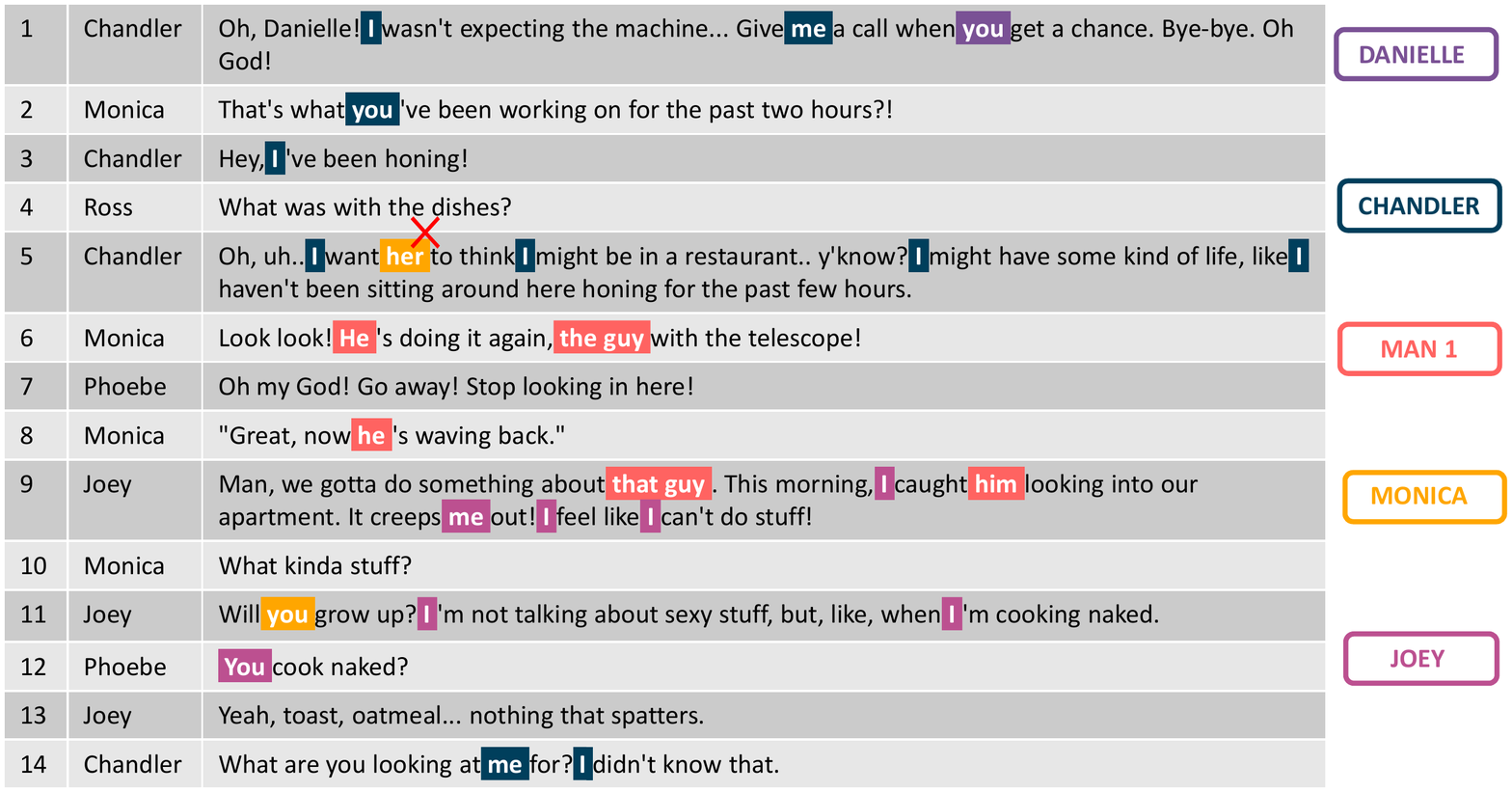}
\end{center}
\caption{
Case study. All mentions that are linked to the same character and in the same coreference cluster are highlighted with the same color. The misclassified mention is marked with the red cross.
}
\label{fig:case}
\end{figure*}

Another contribution of the proposed C$^2$ model is the proposed mention-level self-attention (MLSA) module, which helps iteratively refine the mention representations according to the other mentions co-occurred within the same document.
In this section, to show its effect and the influence of iteration layers, we tried different layers and show their performances on the test set in Figure~\ref{fig:layergraph}.
We conducted the experiments with the SpanBERT-Base encoder and all other hyper-parameters are the same.
The x-axis is the number of layers, and the y-axes are F1 scores of B3, CEAF, and BLANC for coreference resolution, the Macro and Micro F1 scores for character linking.
From the results, we can see that with the increase of layer number from zero to five, the F1 scores on both tasks gradually increase. 
This trend demonstrates that the model can perform better on both tasks when there are more layers. Meanwhile, the marginal performance improvement of the MLSA layer is decreasing.
This indicates that adding too many layers of MLSA may not further help improve the performance because enough context has been included.
Considering the balance between performance and computational efficiency, we chose the iteration layers to be two in our current model based on similar observations made on the development set.

\begin{table}[t]
\begin{center}
\scriptsize
\begin{sc}
\begin{tabular}{p{1.3cm}|C{0.8cm}C{0.8cm}C{0.8cm}|C{0.8cm}C{0.8cm}}

\toprule
 & \multicolumn{3}{c|}{Coreference F1} & \multicolumn{2}{c}{Linking F1}   \\ \midrule

Model & B3 & CEAF{$\phi4$} & BLANC & Micro    & Macro  \\ 
\midrule

C$^2$ &\textbf{85.54}  & \textbf{ 77.48}   & \textbf{92,17}  &   \textbf{87.05}   &  \textbf{81.09}  \\
\hspace{0.1cm} - MLSA & 83.57 & 75.32  & 90.51  &   86.26   &  80.32  \\
\hspace{0.1cm}  - Linking  & 83.50 & 76.10   & 90.08   &  - &   -   \\
 \hspace{0.1cm} - Coref.   & - & -   & -  &   86.94  &  79.58  \\
\bottomrule
\end{tabular}
\end{sc}
\end{center}
\caption{Three ablation studies are conducted concerning the MLSA layers, the coreference resolution module, and the character linking module.}
\label{tab:Ablation}
\end{table}

\subsection{Ablation Study}

In this section, we present the ablation study to clearly show the effect of different modules in the proposed framework C$^2$ in Table~\ref{tab:Ablation}. 
First, we try to remove the mention-level self-attention (MLSA) from our joint learning model and a clear performance drop is observed on both tasks. 
Specifically, the performance on coreference resolution is reduced by 1.21 on the average F1, and meanwhile, the macro-F1 and micro-F1 scores on character linking decreased by 0.77 and 0.79 respectively.
The reduction reveals that the MLSA indeed helps achieve better mention representations with the help from co-occurred mentions. 
Second, we try to remove the coreference resolution and character linking modules.
When the character linking module is removed, it is observed that the performance on coreference resolution decreased by 1.94 on the averaged F1 score.
When the coreference module is removed, the performance of C$^2$ on character linking dropped by 0.83 on the average of Micro and Macro F1 scores. 
These results prove that the modeling of coreference resolution and character linking can indeed help each other and improve the performance significantly, and the proposed joint learning framework can help to achieve that goal.

\subsection{Case Study}

Besides the quantitative evaluation, in this section, we present the case study to qualitatively evaluate the strengths and weaknesses of the proposed C$^2$ model.
As shown in Figure \ref{fig:case}, we randomly select an example from the development set to show the prediction results of the proposed model on both tasks.
To illustrate the coreference resolution and character linking results from the C$^2$ model, the mentions from the same coreference cluster are highlighted 
with the same color. 
Also, we use the same color to indicate to which character the mentions are referring. 
Meanwhile, the falsely predicted result is marked with a red cross.

\subsubsection{Strengths}
For this example, the results on both tasks are consistent. 
The mentions that are linked to the same character entity are in the same coreference group and vice versa. 
Based on this observation and previous experimental results, it is more convincing that the proposed model can effectively solve the two problems at the same time.
Besides that, we also notice that the model does not overfit the popular characters. 
It can correctly solve all the mentions referring to not only main characters, and also for the characters that only appear several times such as \textit{MAN 1}. 
Last but not least, the proposed model can correctly resolve the mention to the correct antecedent even though there is a long distance between them in the conversation.
For example, the mention \textit{me} in utterance 14 can be correctly assigned to the mention \textit{you} in utterance 2, though there are 11 utterances in between.
It shows that by putting two tasks together, the proposed model can better utilize the whole conversation context.
The only error made by the model is incorrectly classifying a mention and at the same time putting it into a wrong coreference cluster.

\subsubsection{Weaknesses}
By analyzing the error case, it is noticed that the model may have trouble in handling the mentions that require common sense knowledge.
Humans can successfully resolve the mention \textit{her} to Danielle because they know Danielle is on the other side of the telephone, but Monica is in the house. 
As a result, Chandler can only deceive Danielle but not Monica. 
But the current model, which only relies on the context, cannot tell the difference. 

\subsection{Error Analysis}
We use the example in Figure \ref{fig:case} to emphasize the error analysis that compares the performance of our model and the baseline models. The details are as follows. In this example, the only mistake made by our model is related to common-sense knowledge, and the baseline models are also not able to make a correct prediction. 
 
For coreference resolution, 3 out of 25 mentions are put into a wrong cluster by the c2f baseline model. The baseline model failed to do long-distance antecedent assignments (e.g., the “me” in utterance 14).  Meanwhile, our model is better in this case because it successfully predicts the antecedent of the mention “me”, even though its corresponding antecedent is far away in utterance 2. This example demonstrates the advantage that our joint model can use global information obtained from character linking to better resolve the co-referents that are far away from each other.
 
For character linking, 2 out of 25 mentions are linked to the wrong characters by the baseline model. It is observed that the baseline model cannot consistently make correct linking predictions to less-appeared characters, for example, the “He” in utterance 6. 
In this case, our model performs better mainly because it can use the information gathered from the nearby co-referents to adjust its linking prediction,  as its nearby co-referents are correctly linked to corresponding entities.

\section{Related Works}

Coreference resolution is the task of grouping mentions to clusters such that all the mentions in the same cluster refer to the same real-world entity~\cite{Pradhan2012CoNLL,DBLP:conf/naacl/ZhangSS19,DBLP:conf/acl/ZhangSSY19,DBLP:conf/emnlp/YuZSSZ19}.
With the help of higher-order coreference resolution mechanism~\cite{lee-etal-2018-higher} and strong pre-trained language models (e.g., SpanBERT~\cite{joshi2019coref}), the end-to-end based coreference resolution systems have been achieving impressive performance on the standard evaluation dataset~\cite{Pradhan2012CoNLL}.
Recently, motivated by the success of the transfer learning, \citet{wu2019coreference} propose to model the coreference resolution task as a question answering problem.
Through the careful fine-tuning on a high-quality QA dataset (i.e., SQUAD-2.0~\cite{DBLP:conf/acl/RajpurkarJL18}), it achieves the state-of-the-art performance on the standard evaluation benchmark.
However, as disclosed by~\citet{DBLP:journals/corr/abs-2009-12721}, current systems are still not perfect. 
For example, they still cannot effectively handle pronouns, especially those in informal language usage scenarios like conversations.
In this paper, we propose to leverage the out-of-context character information to help resolve the coreference relations with a joint learning model, which has been proven effective in the experiments.

As a traditional NLP task, entity linking~\cite{10.1145/1321440.1321475,Ji2015OverviewOT,  kolitsas2018endtoend, raiman2018deeptype, onando2020evaluating, van_Hulst_2020} aims at linking mentions in context to entities in the real world (typically in the format of knowledge graph).
Typically, the mentions are named entities and the main challenge is the disambiguation.
However, as a special case of the entity linking, the character linking task has its challenge that the majority of the mentions are pronouns.
In the experiments, we have demonstrated that when the local context is not enough, the richer context information provided by the coreference clusters could be very helpful for linking mentions to the correct characters.

In the NLP community, people have long been thinking that the coreference resolution task and entity linking should be able to help each other.
For example, \citet{ratinov-roth-2012-learning} show how to use knowledge from named-entity linking to improve the coreference resolution, but do not consider doing it in a joint learning approach.
After that, \citet{hajishirzi-etal-2013-joint} demonstrate that the coreference resolution and entity linking are complementary in terms of reducing the errors in both tasks. 
Motivated by these observations, a joint model for coreference, typing, and linking is proposed \cite{durrett2014joint} to improve the performance on three tasks at the same time.
Compared with previous works, the main contributions of this paper are two-fold: (1) we tackle the challenging character linking problem; (2) we design a novel mention representation encoding method, which has been shown effective on both the coreference resolution and character linking tasks.

\section{Conclusion}

In this paper, we propose to solve the coreference resolution and character linking tasks jointly.
The experimental results show that the proposed model C$^2$ performs better than all previous models on both tasks.
Detailed analysis is also conducted to show the contribution of different modules and the effect of the hyper-parameter.

\section{Acknowledgements}
This paper was supported by the NSFC Grant U20B2053 from China, the Early Career Scheme (ECS, No. 26206717),  the General Research Fund (GRF, No. 16211520), and the Research Impact Fund (RIF, No. R6020-19) from the Research Grants Council (RGC) of Hong Kong, with special thanks to the Tencent AI  Lab  Rhino-Bird  Focused  Research  Program.

\bibliography{eacl2021}
\bibliographystyle{acl_natbib}

\end{document}